\pdfoutput=1

\documentclass[letterpaper, 10 pt, conference]{IEEEtran}  

\IEEEoverridecommandlockouts                              

\usepackage{epsfig} 
\usepackage{mathptmx} 
\usepackage{times} 
\usepackage{amsmath} 
\usepackage{amssymb}  

\usepackage{graphicx} 
\usepackage{graphics} 
\usepackage{grffile} 
\usepackage[T1]{fontenc}
\usepackage{amsfonts}
\usepackage{booktabs}
\usepackage{siunitx}
\usepackage{caption} 
\usepackage{booktabs}
\usepackage{graphicx}
\usepackage{algorithmic}
\usepackage[ruled]{algorithm2e}
\usepackage{float}

\usepackage{mwe,tikz}\usepackage[percent]{overpic}

\usepackage{subcaption}
\usepackage{array,   tabularx, makecell, booktabs}%
\usepackage{geometry}
\usepackage{comment}

\DeclareMathOperator*{\argmax}{arg\,max}
\DeclareMathOperator*{\argmin}{arg\,min}
\title{\LARGE \bf
Enhancing Navigational Safety in Crowded Environments using Semantic-Deep-Reinforcement-Learning-based Navigation
}

\author{Linh K{\"a}stner$^{1}$\thanks{$^{1}$Linh K{\"a}stner, Junhui Li, Zhengcheng Shen and Jens Lambrecht are with the Chair Industry Grade Networks and Clouds, Faculty of Electrical Engineering, and Computer Science,				
		Berlin Institute of Technology, Berlin, Germany
		{\tt\small linhdoan@tu-berlin.de}}, Junhui Li$^{1}$, Zhengcheng Shen$^{1}$, and Jens Lambrecht$^{1}$
}

\begin{document}

\maketitle
\thispagestyle{empty}
\pagestyle{empty}


\begin{abstract}
\noindent Intelligent navigation among social crowds is an essential aspect of mobile robotics for applications such as delivery, health care, or assistance. Deep Reinforcement Learning emerged as an alternative planning method to conservative approaches and promises more efficient and flexible navigation. However, in highly dynamic environments employing different kinds of obstacle classes, safe navigation still presents a grand challenge. In this paper, we propose a semantic Deep-reinforcement-learning-based navigation approach that teaches object-specific safety rules by considering high-level obstacle information. In particular, the agent learns object-specific behavior by contemplating the specific danger zones to enhance safety around vulnerable object classes. We tested the approach against a benchmark obstacle avoidance approach and found an increase in safety. Furthermore, we demonstrate that the agent could learn to navigate more safely by keeping an individual safety distance dependent on the semantic information. 

\end{abstract}

\section{Introduction}
Mobile robots have gained significant importance due to their flexibility and the variety of use cases in which they can operate \cite{fragapane2020increasing}, \cite{alatise2020review}. Whereas classic planning approaches can cope well with static environments, reliable obstacle avoidance in dynamic environments remains a central challenge. Navigation among crowds is especially complex due to different types of humans, each with different behavior, which in turn demand different navigation strategies. Reliable and safe navigation in these highly dynamic environments is essential to the operation of mobile robotics \cite{robla2017working}. 
\begin{figure}[!h]
	\centering
	\includegraphics[width=2.4in, height=2.2in]{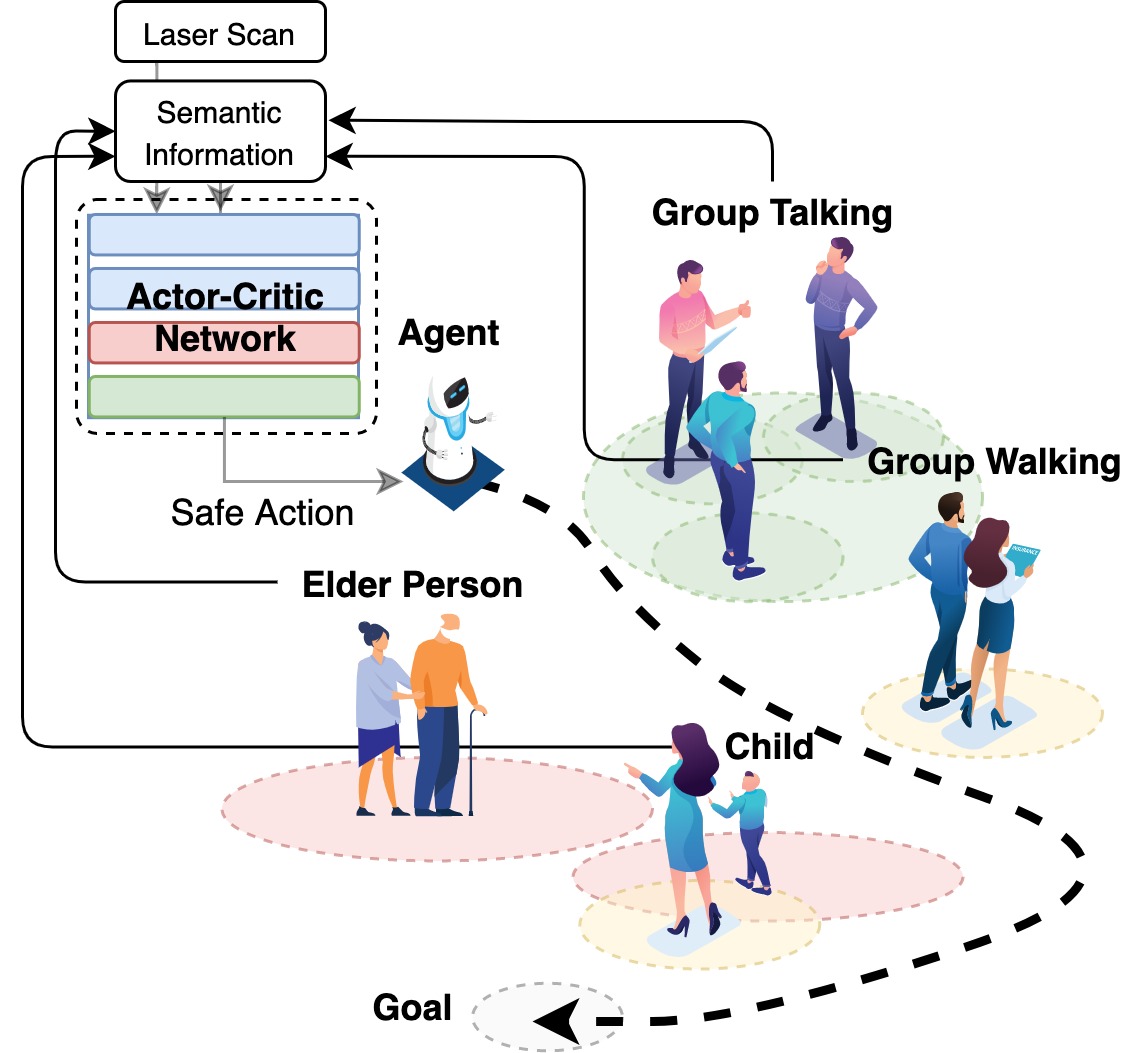}
	\caption{This work proposes a semantic DRL-based navigation approach, which adds semantic information about nearby obstacles to the DRL pipeline. The agent should learn object-specific behavior and move more safely around vulnerable object classes like elders and children. While previous research work proposed efficient ways to navigate within crowds, our work emphasizes navigational safety.}
	\label{intro}
\end{figure}
Deep Reinforcement Learning (DRL) emerged as an end-to-end learning approach, which directly maps the raw sensor outputs to robot actions and has shown promising results in teaching complex behavior rules, increasing robustness to noise, and generalizing new problem instances. In recent years, a variety of literature has incorporated DRL into robot navigation with remarkable results \cite{chiang2019learning},\cite{chen2017socially}. 
However, most approaches explicitly trained the agent to avoid obstacles in an efficient manner, which could result in the agent being too proximate to the obstacles, and thus not adhering to safety measures. This is especially problematic in environments where the robot has to react and adapt its behavior to specific situations and pay additional attention to vulnerable objects like elders and children. In this paper, we propose a semantic DRL-based navigation approach, where an agent learns object-specific safety rules by considering high-level information about obstacle classes to enhance safety among crowds. More specifically, we realistically model different obstacle classes within a 2D simulation and train the agent to adapt its behavior accordingly. While previous research proposes efficient ways to navigate within crowds, our work emphasizes navigational safety. The main contributions of our paper are the following:

\begin{itemize}
    \item Extension of our previous work, arena-rosnav \cite{kastner2021towards}, with multiple different obstacle classes that are modeled realistically using the social force model of Helbing et al. \cite{helbing1995social} 
    \item Modeling of two safety zones and integration into the DRL pipeline. Therefore, we formulate semantic reward systems to shape the agent's behavior and propose two semantic DRL-based planners
    \item Evaluation of navigational safety against a baseline DRL planner within highly complex, collaborative environments consisting of three different types of obstacles: adult, elder, child.
\end{itemize}
The paper is structured as follows: Sec. II begins with related works, followed by the methodology in Sec. III. Subsequently, the results and evaluations are presented in Sec IV. Finally, Sec. V provides a conclusion and outlook.
We made our code open-source at https://github.com/ignc-research/navsafe-arena.


\section{Related Works}
Autonomous navigation of mobile robots has been extensively studied in various research publications. While current navigation systems work well in static environments or dynamic environments, navigation in crowds is especially complex due to the volatile behavior patterns of humans.
Typically, model-based methods such as TEB \cite{rosmann2015timed} or MPC \cite{rosmann2019time} or reactive methods like RVO \cite{van2008reciprocal} or OCRA \cite{van2011reciprocal} are being developed for obstacle avoidance in crowded environments. These methods work well under certain assumptions but require well-engineered parameters and rules. 
Moreover, in collaborative environments employing various behavior classes, these hand-engineered rules could become tedious and computationally intensive. 
Deep Reinforcement Learning (DRL) has emerged as an end-to-end approach with the potential to learn collaborative behavior in unknown environments. Various publications utilize DRL for socially-aware obstacle avoidance \cite{chiang2019learning}, \cite{faust2018prm}, \cite{francis2020long}. To train navigation approaches in realistic crowds, a variety of research works leverage the social force model \cite{helbing1995social} to provide realistic training platforms. 
Fan et al. \cite{fan2018crowdmove} proposed CrowdMove, a DRL-based collision avoidance approach trained in a 2D simulation environment based solely on laser scan data. The researchers deployed the approach on four different robotic platforms and demonstrated robust obstacle avoidance. Similarly, Liang et al. \cite{liang2020crowd} proposed Crowdsteer, which fuses a 2D lidar with RDB-D data and computes smooth, collision-free trajectories. The researchers demonstrate their approach on a 3D simulator and accomplish robust performance in moderate crowds. 
Chen et al. \cite{chen2019crowd} propose a novel architecture using transformers to encode human interactions to be considered within their simulation environment CrowdNav. In follow-up work \cite{chen2019relational}, the researchers included a graph convolutional network to predict the behavior of humans and collaborative robots. Dugas et al. \cite{dugas2020navrep} proposed a 2D simulation platform to train and test DRL-based agents on highly dynamic human environments.
Guldenring et al. \cite{guldenringlearning} propose a ROS-integrated simulator based on the social force model of Helbing et al. \cite{helbing1995social} and trained several DRL-agents within that environment. Rudenko et al. \cite{inproceedingsRuden} propose to utilize the same social force model to predict future human trajectories. Similarly, Sun et al. \cite{sun2020inverse}, Truong et al. \cite{truong2017toward} and Ferrer et al. \cite{ferrer2019anticipative} include a human motion and behavior prediction algorithm into their navigation system to enhance environmental understanding and adjust the navigation for crowded environments. A socially aware DRL-based planner (CADRL) was proposed by Chen et al. \cite{chen2017socially}. They introduced social norms like keeping to the right side of the corridor or driving left for overtaking maneuvers. Everett et al. \cite{everett2018motion} extend the CADRL approach with LSTM architecture. 
Other works that present DRl-based navigation approaches for crowded environments include \cite{liu2020robot} or \cite{choi2019deep}.
Samsani et al. \cite{samsani2021socially} integrated the concept of safety zones from \cite{bera2017sociosense} and \cite{932821} to the CrowdNav environment proposed by Chen et al. \cite{chen2019crowd} to enhance navigational safety.
Of note, all the aforementioned approaches explicitly trained the agent to avoid obstacles in an efficient manner. However, this could result in the agent being too proximate to the obstacles and thus not adhering to safety measures.
In collaborative environments, the robot ideally adapts its behavior to specific kinds of object classes. For instance, the agent should be more careful around children or elders, but maintain its speed around adults. On that account, this work proposes an end-to-end solution where the agent learns to adapt its behavior and navigation style depending on the object class.

\begin{figure*}[!h]
    \centering
		\includegraphics[width= 0.75\textwidth]{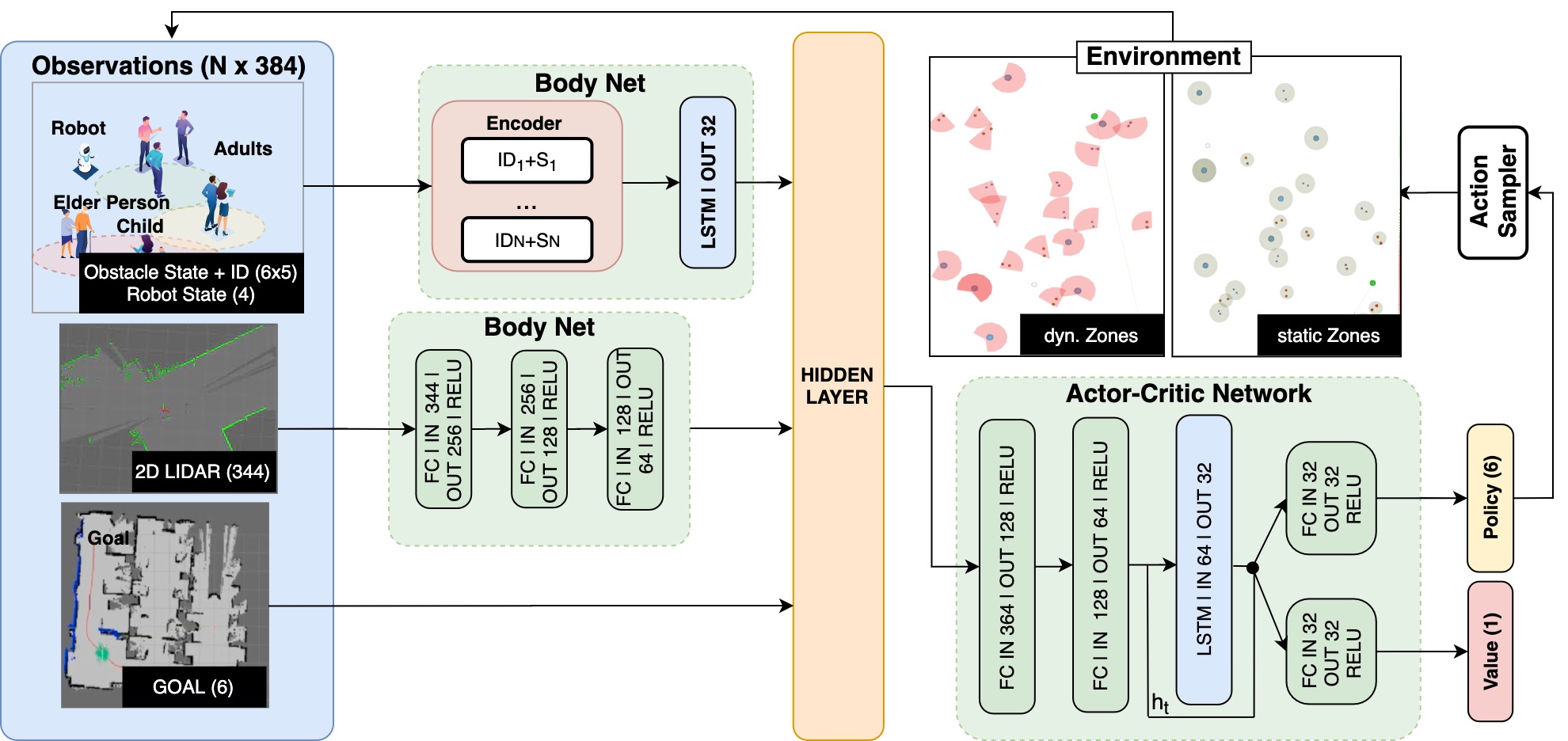}
	\caption{System design. The arena-rosnav platform \cite{kastner2021towards} is extended with three different kinds of obstacles: adults, older adults, and children. It enables the training and testing of semantic-DRL-based approaches with the ROS navigation stack. The semantic information about the obstacle class and the obstacle position is encoded with an LSTM architecture. The sensor observations and the goal position are directly provided as input into the hidden layer. An actor-critic network uses the accumulated input to train object-specific behavior. The training environments are randomized to avoid overfitting and three different training modes are provided: scenario-, random-, and staged training.}
	\label{designsys}
\end{figure*}

\section{Methodology}
In this section, we will present the methodology of our framework for teaching the agent object-specific behavior. The agent should maintain greater proximity from and exercise more caution around vulnerable classes, such as older adults and children. To accomplish this adapted behavior, we employed a DRL-based approach, which uses the obstacle-related semantic information as input.

\subsection{Problem Formulation}
We formalize the robot navigation problem as a partially observable Markov decision problem (POMDP) that can be described by a 6-tuple $(S,A,P,O,R, \gamma)$
where S represents the robot states, A the action space, P the transition probability from one state to another, R the reward, O the observations, and $\gamma$ the discount factor. Next, we formulate the following task constraints for semantic navigation towards the goal:
\begin{align}
   C_{j,i,t}(s,a) = 
   \begin{cases} 
        |p_r - p_g|_2 < d_{goal} &  \textit{ goal constraint}   \\
        \min(O_{L,t}) > d_r  & \mbox{ \textit{collision constraint} } \\
        \min(O_{S,t}) > d_z  & \mbox{ \textit{semantic constraint} } \\
        \argmin [t | s, \pi]  & \mbox{ \textit{time constraint} } \\
    \end{cases}  
\end{align}
Where $d_{goal}$ is the goal's radius, $p_r$ is the robot's position, $p_g$ is the goal's position, O is the robot's observation including the LiDAR observation $O_L$ and the semantic information $O_S = [s_p, s_r]$, $d_r$ is the robot's radius and $d_z$ is the safety zone's radius. $O_l$ is thereby represented by a 2D LiDAR with 360 points in the ring including angles and distances. The semantic observation $O_S = [s_p, s_r]$ consists of two parts. The first part $s_p$ is formulated as the state of every nearby human in the vicinity of the robot including the position $(\tilde{p_x},\tilde{p_y})$, the size $r_h$, the size of the safety zone $r_z$, the distance to the agent to the human $d_{ah}$, and the obstacle type ID. The second part is the state of the robot $s_r$ including the distance to the goal $d_{ag}$, the position $(p_x,p_y)$, the velocity $(v_{linear},v_{angular})$, and the radius of the robot $r_a$:
\begin{align}
    s_p=[\tilde{p_x},\tilde{p_y},r_h,d_{ah},r_z,r_a+r_z, id]
\end{align}
\begin{align}
    s_r=[d_{ag},p_x,p_y,v_{linear},v_{angular},r_a]
\end{align}

\subsection{Neural Network Architecture}
The system design and neural network architecture are illustrated in Fig. \ref{designsys}.
The agent is provided with the LiDAR perception and semantic information as input to not only learn to navigate safely around specific obstacles autonomously but also adjust its behavior depending on the encountered obstacle classes. The semantic information about the obstacles are assumed to be perfect. In real-world settings, this obstacle information could be obtained by state-of-the-art computer vision approaches using cameras. 
The two different streams of information, the semantic input, and the laserscan are processed separately. The obstacle ID, the respective obstacle state as well as the robot state are first provided as input into an LSTM network to encode the interaction between each human and robot. The LiDAR scan is provided into 3 fully connected layers. Subsequently, these streams are merged together with the global goal position and the time into the hidden layer. Finally, the vector is given as input into the actor-critic network. We utilize an asynchronous policy-gradient method, which directly learns the policy based on proximal policy optimization as described by Schulman et al. \cite{schulman2017proximal}. The output is split into a policy (actor) and a value (critic) function where the value function assesses the decisions made by the policy function. Subsequently, the optimal policy can be adjusted using policy iteration \cite{sutton2018reinforcement}. We train on a continuous action space for more flexibility and smoothness of actions  \cite{faust2018prm}. The action space consists of a forward velocity $v_{linear} \in [0,0.6] \quad m/s$ and a rotational velocity $v_{angular} \in [-\pi/6,\pi/6] \quad rad/s $.

\subsection{Behavior Modeling - Object Models, Danger Zones}
\begin{figure}[!h]
    \centering
	\includegraphics[width=0.29\textwidth]{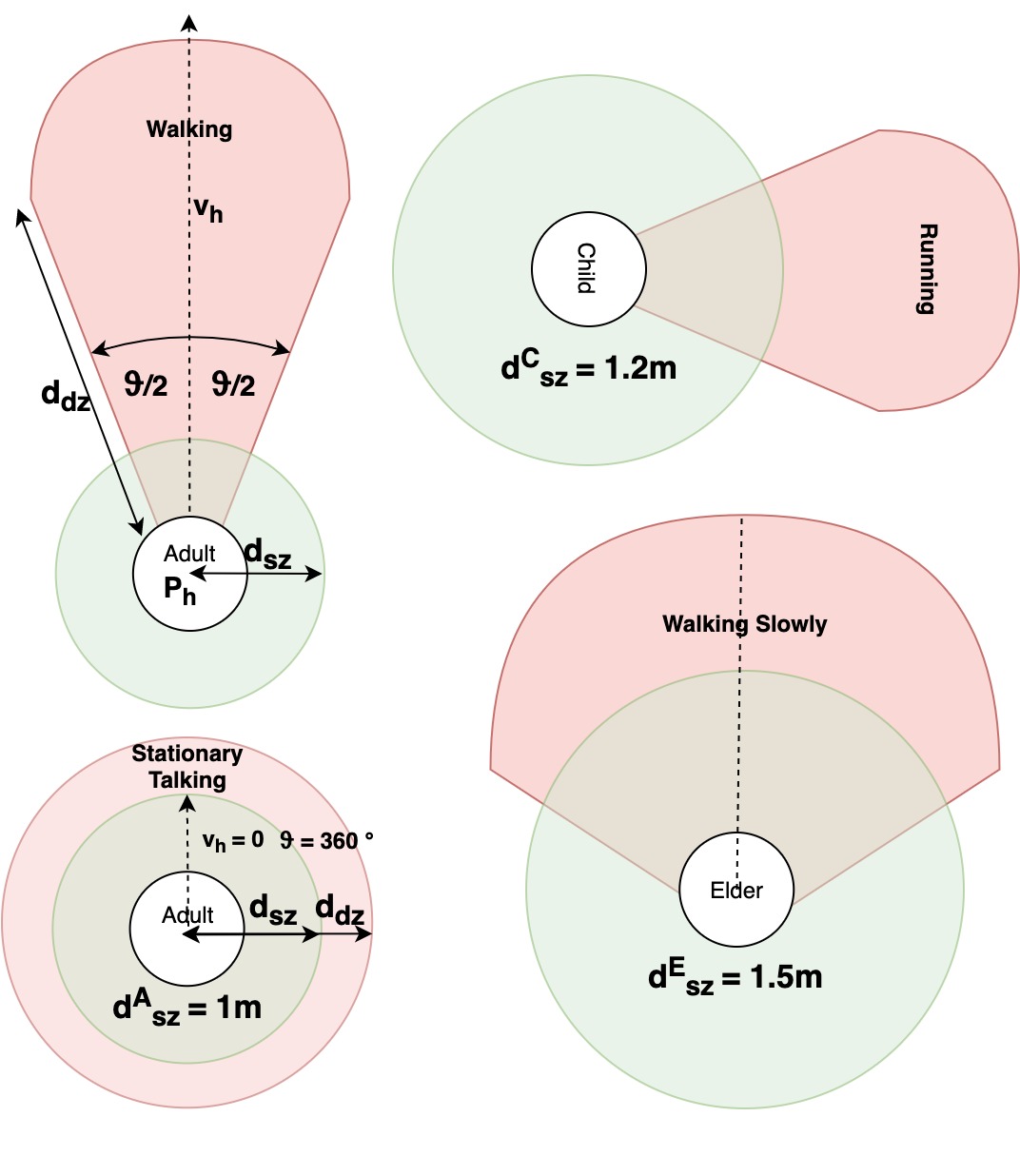}
	\caption{Object behavior modeling using two safety zone models. The static safety zone is implemented using a static, object-specific radius, while the dynmic safety zone is calculated based on the work of Samsani et al. \cite{samsani2021socially}.}
	\label{zones}
\end{figure}

To shape the agent's behavior around specific types of obstacles, we model object-specific safety zones.
Typically, there exist different types of obstacles each with its own behavior. We integrate three different object classes into our simulator arena-rosnav \cite{kastner2021towards} using the Pedsim model \cite{helbing1995social} to emulate realistic behavior patterns: child, adult, older adult. Whereas the adult will move at an average speed of 0.6 m/s, the children and elderly move at a slower pace of 0.4 and 0.1 m/s respectively. Note that the specific selection of these different object types was based on the applicability to common use cases such as navigation in airports or train stations, but can be extended to different object classes and preferences e.g. for industrial setups with forklifts, different vehicle types, or robots. To model these safety zones, two modeling approaches are implemented: (1) static safety zones, which contain object-specific safety zones of a fixed size for each individual obstacle, (2) dynamic safety zones based on the velocity of the obstacle. 
The static safety zones are implemented with a static object-specific radius $d_{sz}$ around the obstacles, which are modeled as follows: $d^A_{sz} = 1m$ for adult, $d^C_{sz} = 1.2m$ for the child, and $d^E_{sz} = 1.5m$ for elder.
The dynamic safety zones are implemented based on \cite{samsani2021socially} by calculating the zones depending on the obstacle velocity $v_h$:

\begin{align}
    d_{dz} = k_v \cdot v_h + r_{static} \\
    \vartheta = \frac{11 \pi}{6} e^{-1.4 a_v \cdot v} + \frac{\pi}{6}
\end{align}
Here, $d_{dz}$ is the safety zone length, which increases with the velocity of the object and $\vartheta$ is the angle of the safety zone, which decreases exponentially with the velocity of the object. $r_{static}$ is equal to $d^A_{sz}$ for adult, $d^C_{sz}$ for child and $d^E_{sz}$ for elder.   The factor $k_v$ is set to 1.5 as proposed by \cite{samsani2021socially}. Different to \cite{samsani2021socially}, we add the factor $a_v=1.5$ for a more realistic decrease of the zone. The safety zones are illustrated in Fig. \ref{zones}. 

Subsequently, these models are used to formalize our reward functions.

\subsection{Reward systems}
To shape the agent's behavior, two specific reward systems for each respective modeling approach are implemented.
The two reward systems share the basic sub-rewards, which are denoted as $R_{base}$:
\begin{align}
    R_{base}(s_t,a_t) = [r_{s}^t + r_{c}^t + r_{p}^t ]
\end{align}
Firstly, the robot will obtain a reward $r_{arrival}$ if it reaches the goal. Here, we set $r_{arrival} = 2$ and $p_r$ is the position of the robot and $p_g$ is the position of goal. 
\begin{align}
r_{s}^t&=\begin{cases}r_{arrival}& \textit{if }|p_r - p_g|_2 < d_{goal}\\0&otherw.\end{cases}
\end{align}
Correspondingly, the robot is penalized if a collision happens. The collisions are detected by using the laser scan. Here, $O_{L,t}$ represents the observation of the 2D LiDAR, $d_r$ is the radius of the robot. We set $r_{collision} = -4$.
\begin{align}
r_{c}^t&=\begin{cases}r_{collision}&\textit{if } \min(O_{L,t}) < d_r\\ 0& otherw.\end{cases}
\end{align}
To encourage the robot to keep approaching the goal, a progress reward $r_p$ is given.
\begin{align}
r_{p}^t& =\begin{cases}w_{p}(t) & d^t>0,\;d^t=d_{ag}^{ t-1}-d_{ag}^{ t}\\
w_{s} & \text {else if } d^t=0 \\ w_{n}&  otherw.\end{cases}
\end{align}
Here, $d_{ag}$ denotes the distance between the agent and the goal and the difference between the current time step and the last time step, $d_{ag}=|p_r - p_g|_2$, is calculated to monitor the progress. We set $w_{p}(t) = 0.018e^{1-t}, w_{s} =-0.03, w_{n} = -0.14$. The reward will reduce along with the time in order to urge the robot to find the optimal path in a reasonable time. 
Subsequently, we define the static zone (SZ) reward $r_{sz}$ to model the safety zones for each object as follows: 
\begin{align}
r_{sz}^t&=\begin{cases}k_{sz}*e^{1-d_{ah}/r_{sz}} & \text {if exceed } \mathrm{SZ}\\0&otherw.\end{cases}
\label{r1}
\end{align}
Here, $d_{ah}$ denotes the distance between the agent and the human, while $r_{sz}$ denotes the radius of each static safety zone SZ, which differs according to the type of human. 
As inferred from the formula, the robot will be penalized when exceeding the static zone, that is $d_{ah} < r_{sz}$, the exponential penalty is controlled by a factor $k_{sz} = -0.08$.
The dynamic zone (DZ) reward $r_{dz}$ is designed as following:

\begin{align}
r_{dz}^t&=\begin{cases}-\left(d_{c}\left(\frac{-0.15}{d_{dz}-R}\right)+0.15\right) & \text {if in } \mathrm{DZ}\\0&otherw.\end{cases}
\label{r2}
\end{align}
Here, $d_{c}$ denotes the distance between the agent and the human minus both radiuses. $d_{dz}$ denotes the radius of the dynamic danger zone, $R$ denotes the human radius,  which differs according to the type of human. Similarly, if the danger zone is exceeded ($d_{ag}-R-d_r < d_c$), the agent is penalized. 
Given the safety distance sub-reward functions in Eq. \ref{r1} and \ref{r2}, our two reward systems are defined as follows:
\begin{align}
    R_{static}(s_t,a_t) &= R_{base}(s_t,a_t) +  r_{sz}^t \\
    R_{dynamic}(s_t,a_t) &= R_{base}(s_t,a_t) +   r_{dz}^t 
\end{align}

\begin{figure*}[!h]
	
	\begin{subfigure}{0.3\textwidth}(a)
	 \centering 
		\includegraphics[width=\linewidth]{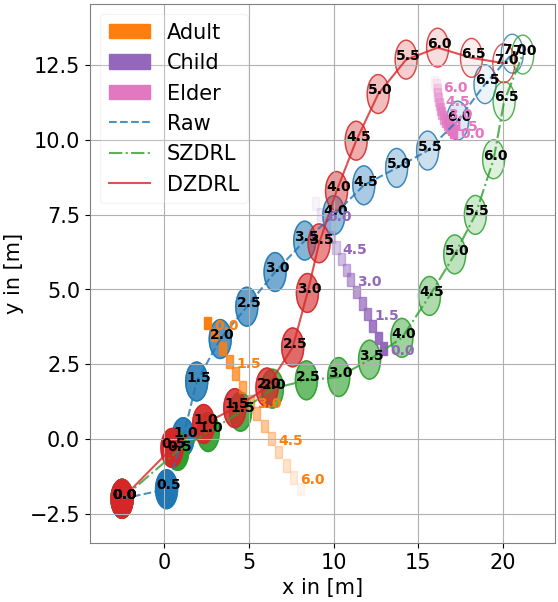}
		
		\label{fig:1}
	\end{subfigure}\hfil 
	\begin{subfigure}{0.3\textwidth }(b)
	 \centering 
		\includegraphics[width=\linewidth]{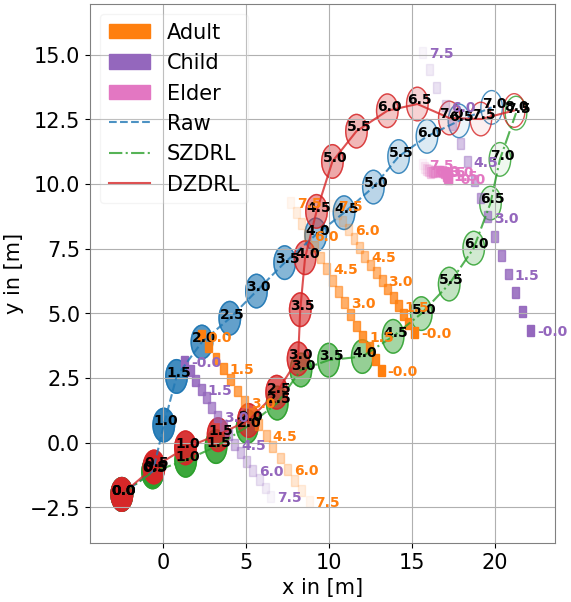}
		
		\label{fig:2}
	\end{subfigure}\hfil 
	\begin{subfigure}{0.3\textwidth}(c)
	 \centering 
		\includegraphics[width=\linewidth]{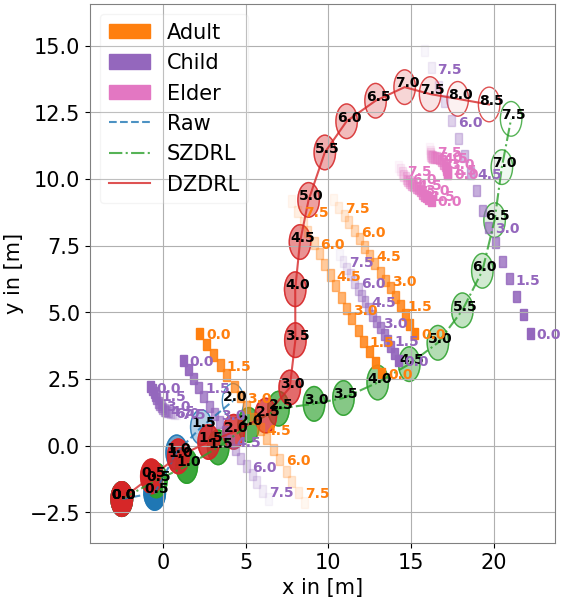}
		
		\label{fig:3}
	\end{subfigure}
	
	\medskip
		\begin{subfigure}{0.28\textwidth}(d)
		 \centering 
		\includegraphics[width=\linewidth]{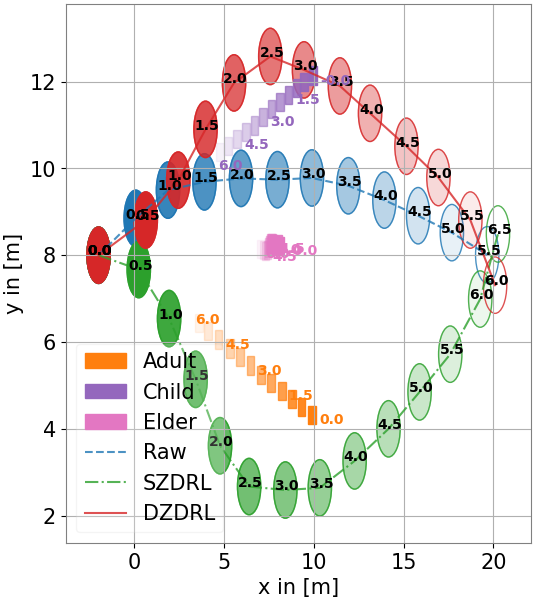}
		
		\label{fig:1}
	\end{subfigure}\hfil 
	\begin{subfigure}{0.28\textwidth }(e)
	 \centering 
		\includegraphics[width=\linewidth]{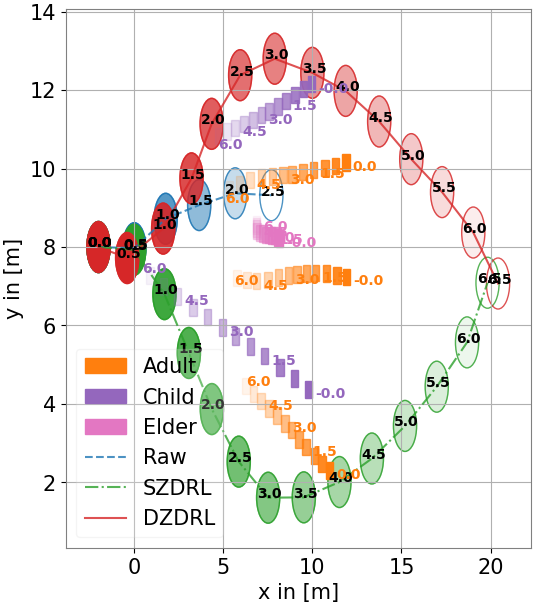}
		
		\label{fig:2}
	\end{subfigure}\hfil 
	\begin{subfigure}{0.28\textwidth}(f)
	 \centering 
		\includegraphics[width=\linewidth]{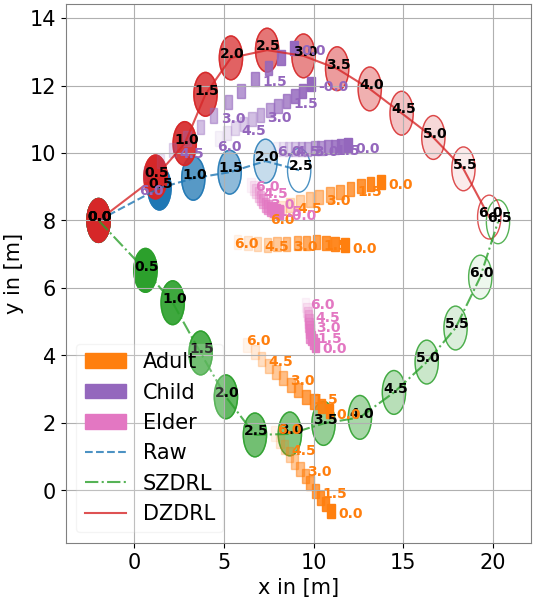}
		
		\label{fig:3}
	\end{subfigure}
	
	\caption{Trajectories of all planners on different test scenarios with adult, child, and elder. (a-c) 3,6, and 9 dynamic obstacles approaching robot from the side, (d-f) 3,6, and 9 dynamic obstacles approaching from the front. Each trajectory is marked with timesteps indicating the position at every timestep t.}
	\label{quali}
\end{figure*}

\subsection{Training Setup}
The agent is trained on randomized environments, which are depicted in Fig. \ref{designsys}. The obstacles are spawned randomly after each episode according to the social force model. This way, we mitigate over-fitting issues and enhance
generalization. Curriculum training is adapted, which spawns
more obstacles once a success threshold is reached and fewer
obstacles if the agent’s success rate is low. The threshold is
defined as the sum over the mean reward. The training was
optimized with GPU usage and trained on an NVIDIA RTX
3090 GPU. It took between 30h and 60h to converge.

\section{Results and Evaluation}
In this chapter, we present the experimental setup, the results of the planners, and the respective evaluations. The agent is trained with randomly placed obstacles and validated on different specific scenarios. Localization of the pedestrians, as well as the pedestrian state and IDs, are assumed to be perfectly known. In a real-world setup, this information could be acquired using cameras and computer vision approaches. As the baseline, we use the DRL approach of our previous work \cite{kastner2021towards}, which only requires 2D sensor observations as input. The baseline with no additional semantic information is denoted as the Raw approach, the one equipped with the static zone is denoted as SZDRL, and the one using the dynamic zone is denoted as DZDRL.

\subsection{Qualitative evaluation}
We tested all approaches in 6 different scenarios with a fixed start and goal position and three to nine obstacles interfering horizontally and vertically. The qualitative trajectories of all planners on each scenario are illustrated in Fig. \ref{quali}. The timesteps are sampled every 0.5s and visualized within the trajectory of all approaches. The trajectories of the different kinds of humans are marked with the timesteps sampled every 1.5s. The episode ended once a collision occurred. 
It is observed that our semantic DRL approaches SZDRL and DZDRL avoid all obstacles with a large margin, whereas the raw DRL approach without safety zones drives more carelessly, resulting in collisions in the scenarios with more than 6 obstacles at (c) 2.0s, (e) 2.5s, (f) 2.5s.
\begin{table*}[!h]
	\centering
	\setlength{\tabcolsep}{2pt}
	\renewcommand{\arraystretch}{1}

	\caption{Table for quantitative evaluation results}
\begin{tabular}{ll|p{1.2cm}p{1.2cm}p{1.2cm}p{1.4cm}|p{1.0cm}p{1.0cm}p{1.0cm}p{1.0cm}|p{1.0cm}p{1.0cm}p{1.0cm}p{1.0cm}}
\toprule
              
               \midrule
& Planner &   Time [s]      &    Path [m]   &      Col.  [\%]   &     Success [\%]    
          &    $d_a$  [m]     &   $d_c$  [m]    &   $d_e$   [m]          &  $d_{avg}  [m]$
          &    $t_a$  [s]     &  $t_c$ [s]   &    $t_e$   [s]        &  $t_{avg}$ [s]\\
\midrule
        
                & raw DRL &   \textbf{17.45} & \textbf{14.84} &        26.25 &   70.75 &  1.49 &  1.58 &        1.53 &   1.53 &  1.01 &  1.85 &        3.07 &    1.98 \\
               & SZDRL &   19.08 & 16.32 &        8.25 &   88.0 &  \textbf{1.70} &  1.87 &        \textbf{2.01} &   \textbf{1.86} &  \textbf{0.50} & \textbf{0.92} &        \textbf{1.13}&    \textbf{0.85} \\
               & DZDRL &   22.31 & 17.64 &       \textbf{6.25} &   \textbf{93.75} &  1.66 &  \textbf{1.88} &        1.97 &   1.84 &  0.83 &  1.50 &        1.54 &    1.29 \\

\bottomrule
\end{tabular}
\label{quanttable4}
\end{table*}
\noindent Notably, both our DRL planners with semantic information act more carefully around vulnerable object classes like child and elder, waiting for the human to pass or keep a larger distance around them, whereas the raw approach does not consider any additional information about the obstacle and treats each obstacle equally.
For instance, in scenario (b) 1.5 s, it is observed that both our semantic planners wait for the class to move away, whereas the raw mode passes in front of the child and elder. This often results in a collision e.g. in scenario (c) at 2.0 s where both our semantic planners wait for the child and elder to pass before they proceed to move. 
Additionally, the semantic planners are able to keep different safety distances for the different obstacle types. In scenario (a) at 5.5s, DZDRL drives upwards to create a larger distance to the approaching elder at position (x:17, y:11). This behavior can also be observed in (b) or (c) at 4.5s and 5.0s, respectively. Whereas these planners keep a large safety distance to elders and children, they keep a small distance to adults similar to the raw approach. For instance, in scenario (a) at 2.0s, both semantic approaches only keep a small distance around the adult at position (x:3m,y:4m). 
For the scenarios with vertical obstacle interference (d-f), similar behavior is observed. Notably, SZDRL tends to keep the largest distance to obstacles (green). In all scenarios, SZDRL drives behind the obstacles, whereas the other two planners approached the obstacles from the front, which leads to collisions for the raw approach and more evasive maneuvers for DZDRL. 
As extensively evaluated in our previous work \cite{kastner2021towards}, the collision rates of the raw DRL approaches are already significantly lower compared to those of model-based approaches like TEB \cite{rosmann2015timed} or MPC \cite{rosmann2019time} because they can react to incoming obstacles more efficiently by directly processing the sensor observations. These approaches, however, tend to focus on efficient navigation and navigate very closely to all obstacles due to the lack of semantic information. On the other hand, our semantic DRL approaches can leverage the additional information about the obstacle type and perform much safer and human-specific navigation while maintaining efficiency and robustness. The navigation behavior of our planners is demonstrated visually in the supplementary video.

\subsection{Quantitative Evaluations}
To quantitatively evaluate our approach, all agents were tested on random scenarios where the start and goal positions were randomly set and the agent was run 500 episodes each. Subsequently, we calculated the average distance traveled, the average time to reach the goal, the collision rate, and the success rate. Furthermore, we computed the average exceeding time of safety zones with all objects $t_{object}$, which indicates the time where the robot is within the critical safety zone, as well as the average distance to all objects $d_{object}$. The quantitative results are listed in Table \ref{quanttable4}.

\subsubsection{Navigation Performance}
The success rate for the DZDRL approach is the highest with over 93.75 \% success on all scenarios, followed by SZDRL with 88 \% average success. On the other hand, the raw DRL approach only reached a success rate of 70.75 \%. As expected, the semantic DRL approaches are superior in terms of safety with under 8.25 \% average collisions compared to 26.25 \% collisions for the raw DRL approach. In terms of efficiency, the raw DRL approach achieves shorter trajectories towards the goal with an average of 14.84m compared to 16.32m and 17.64m for SZDRL and DZDRL respectively. The raw approach also requires less time to reach the goal with 17.45s compared to both semantic-based approaches. However, this lead is only slight compared to DZDRL with 22.31s. We can conclude that the raw DRL approach is slightly more efficient than semantic DRL approaches, achieves shorter trajectories, and requires less time. However, in terms of collisions, both semantic DRL approaches outperform the raw planners, accomplishing substantially fewer collisions while maintaining competitive efficiency. More importantly, the raw planner treats different classes almost equally. ON the contrary, the SZDRL and DZDRL adapt their navigation behavior for different types of humans to keep different distances.

\subsubsection{Navigational Safety}
\begin{figure}[!h]
    \centering
	\includegraphics[width=0.38 \textwidth]{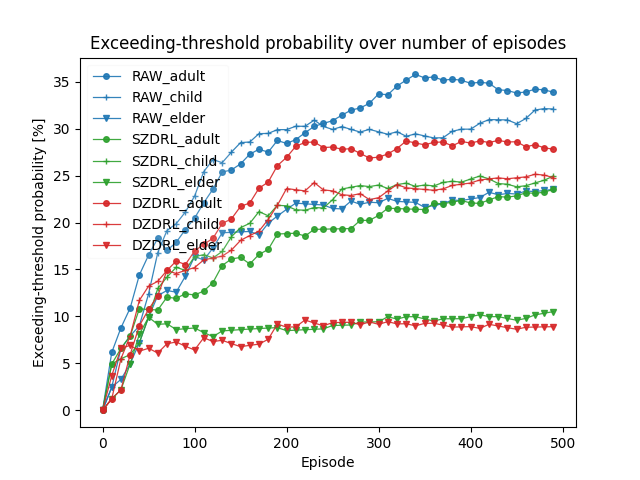}
	\caption{Exceeding-threshold probabilities for all obstacle classes of all planners over 500 episodes.
	}
	\label{probability exceeding1}
\end{figure}

To highlight the improved safety of our semantic agents compared to the raw one, we measured the exceeding rate of all planners to each obstacle class and plotted them in Fig. \ref{probability exceeding1}. The plots statistically indicate the rate of exceeding a pre-defined safety threshold of 1.5 meters for every type of human over a run of 500 episodes after which the values converged. It is observed that the raw approach (blue) exceeds the safety threshold in more than 30 percent of the time for the adult and child class, while our SZDRL agent accomplishes under 20 percent. Notably, both our semantic agents exceeded the safety threshold for the elder class in under 10 percent of the runs, whereas the raw approach exceeded it in more than 20 percent. The results indicate that our semantic agents have a significantly lower probability to exceed a safety threshold compared to the raw approach.

\section{Conclusion}
In this paper, we proposed a semantic DRL-based navigation approach for safe navigation in crowded environments. In particular, we integrated semantic information into the DRL training pipeline. The trained agent could successfully learn and adapt object-specific behavior using only the obstacle type and position as additional input alongside the laser scan observations and relative goal position. More specifically, it was able to behave more safely around vulnerable object classes like children and older adults, while maintaining efficient navigation around adults. Subsequently, we evaluated our semantic DRL planners against a baseline DRL planner without semantic information in terms of safety, efficiency, and robustness. Results demonstrate an improvement in navigational safety for our semantic DRL planners while maintaining competitive efficiency, especially in environments with a large number of obstacles. Future work includes the integration of higher-level semantics, such as social states and interaction patterns to learn more sophisticated behavior, such as human-following or robot guidance tasks. Furthermore, the feasibility of the approach in real environments using computer vision approaches to provide the semantic data is to be extensively evaluated. 

\addtolength{\textheight}{-1cm} 




\typeout{}
\bibliographystyle{IEEEtran}
\bibliography{CNNJournal}

\end{document}